\documentclass[11pt,letterpaper,usenames,dvipsnames]{article}
\usepackage{eacl2017}
\usepackage{natbib}
\usepackage[utf8x]{inputenc}
\usepackage{times}
\usepackage{url}
\usepackage{latexsym}
\usepackage{expex}
\usepackage{colortbl}
\usepackage{booktabs}
\usepackage{amsmath}
\usepackage{amssymb}
\usepackage{adjustbox}
\usepackage{tikz}
\usepackage{ifthen}
\usepackage{ellipsis}

\usetikzlibrary{positioning}
\usetikzlibrary{fit}
\usetikzlibrary{shadows}
\usepackage{algorithm2e}
\usepackage{subcaption}
\usepackage{multirow}
\usepackage{ragged2e}
\usepackage{pgfplots}
\pgfplotsset{compat=1.14}
\usepackage{eurosym}
\usepackage{lscape}
\usepackage{enumitem}
\usepackage{rotating,tabularx}
\usepackage{microtype}
\usepackage{xcolor}
\PrerenderUnicode{ä}

\newcommand{\testdata}{Multi30K 2017 test data}
\newcommand{\testshort}{2017 test}

\title{Findings of the Second Shared Task on Multimodal Machine Translation
and Multilingual Image Description}

\author{ 
  Desmond Elliott\\
  School of Informatics \\
  University of Edinburgh\\
  {\tt d.elliott@ed.ac.uk}\\
  \And
  Stella Frank\\
  Centre for Language Evolution\\
  University of Edinburgh\\
  {\tt stella.frank@ed.ac.uk}\\
  \AND Loïc Barrault \and Fethi Bougares\\
  LIUM\\
  University of Le Mans\\
  {\tt first.last@univ-lemans.fr}\\
  \And Lucia Specia \\
  Department of Computer Science\\
  University of Sheffield\\
  {\tt l.specia@sheffield.ac.uk}\\
}

\date{}
\eaclfinalcopy

\begin{document}
\maketitle

\begin{abstract}
We present the results from the second shared task on multimodal machine translation and multilingual image description. Nine teams submitted 19 systems to two tasks. The multimodal translation task, in which the source sentence is supplemented by an image, was extended with a new language (French) and two new test sets. The multilingual image description task was changed such that at test time, only the image is given. Compared to last year, multimodal systems improved, but text-only systems remain competitive.
\end{abstract}
\section{Introduction}

The Shared Task on Multimodal Translation and Multilingual Image Description tackles the problem of generating descriptions of images for languages other than English.
The vast majority of image description research has focused on English-language description due to the abundance of crowdsourced resources \citep{Bernardi2016}. However, there has been a significant amount of recent work on creating multilingual image description datasets in German \citep{ElliottFrankSimaanSpecia2016,Hitschler2016,rajendran2016bridge}, Turkish \citep{unal2016tasviret}, Chinese \citep{li2016adding}, Japanese \citep{miyazaki2016cross,yoshikawa2017stair}, and Dutch \citep{vanMiltenburgElliottVossen2017}. Progress on this problem will be useful for native-language image search, multilingual e-commerce, and audio-described video for visually impaired viewers.

The first empirical results for multimodal translation showed the potential for visual context to improve translation quality \citep{ElliottFrankHasler2015,Hitschler2016}. This was quickly followed by a wider range of work in the first shared task at WMT 2016 \citep{Specia2016}.

The current shared task consists of two subtasks:
\begin{itemize}

\item Task 1: \textbf{Multimodal translation} takes an image with a source language description that is then translated into a target language. The training data consists of parallel sentences with images.

\item Task 2: \textbf{Multilingual image description} takes an image and generates a description in the target language without additional source language information at test time. The training data, however, consists of images with independent descriptions in both source and target languages.

\end{itemize}

The translation task has been extended to include a new language, French. This extension means the Multi30K dataset \citep{ElliottFrankSimaanSpecia2016} is now triple aligned, with English descriptions translated into both German and French.

The description generation task has substantially changed since last year. The main difference is that source language descriptions are no longer observed for test images. This mirrors the real-world scenario in which a target-language speaker wants a description of image that does not already have source language descriptions associated with it. The two subtasks are now more distinct because multilingual image description requires the use of the image (no text-only system is possible because the input contains no text).

Another change for this year is the introduction of two new evaluation datasets: an extension of the existing Multi30K dataset, and a ``teaser'' evaluation dataset with images carefully chosen to contain ambiguities in the source language.

This year we encouraged participants to submit systems using unconstrained data for both tasks. Training on additional out-of-domain data is under-explored for these tasks. We believe this setting will be critical for future real-world improvements, given that the current training datasets are small and expensive to construct.\footnote{All of the data and results are available at \url{http://www.statmt.org/wmt17/multimodal-task.html}}

\section{Tasks \& Datasets}

\subsection{Tasks}

The Multimodal Translation task (Task 1) follows the format of the 2016 Shared Task \citep{Specia2016}. The Multilingual Image Description Task (Task 2) is new this year but it is related to the Crosslingual Image Description task from 2016. The main difference between the Crosslingual Image Description task and the Multilingual Image Description task is the presence of source language descriptions. In last year's Crosslingual Image Description task, the aim was to produce a single target language description, given five source language descriptions and the image. In this year's Multilingual Image Description task,
participants received only an unseen image at test time, without source language descriptions.

\subsection{Datasets}

\begin{table*}
\centering
\renewcommand{\arraystretch}{1.3}
\begin{tabular}{rcccc}
\toprule
& \multicolumn{2}{c}{Training set} & \multicolumn{2}{c}{Development set}  \\
& Images & Sentences & Images & Sentences \\
\midrule
Translation & 29,000 & 29,000 & 1,014 & 1,014\\
Description & 29,000 & 145,000 & 1,014 & 5,070\\
\midrule
& \multicolumn{2}{c}{\testshort} & \multicolumn{2}{c}{COCO}\\
& Images & Sentences & Images & Sentences\\
\midrule
Translation & 1,000 & 1,000 & 461 & 461\\
Description & 1,071 & 5,355 & \multicolumn{2}{c}{---}\\
\bottomrule
\end{tabular}
\caption{Overview of the Multi30K training, development, \testshort, and Ambiguous COCO datasets.}\label{tab:datasets:overview}
\end{table*}

The Multi30K dataset \citep{ElliottFrankSimaanSpecia2016} is the primary dataset for the shared task. It contains 31K images originally described in English \citep{Young2014} with two types of multilingual data: a collection of professionally translated German sentences, and a collection of independently crowdsourced German descriptions. 

\begin{figure}
\begin{center}
\includegraphics[width=0.4\textwidth]{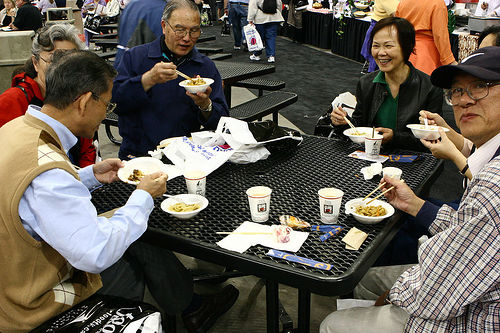}\\
\end{center}
En: A group of people are eating noddles.\\
De: Eine Gruppe von Leuten isst Nudeln.\\
Fr: Un groupe de gens mangent des nouilles.
\caption{Example of an image with a source description in English, together with German and French translations.}\label{fig:examples:en-de-fr}
\end{figure}

This year the Multi30K dataset has been extended with new evaluation data for the Translation and Image Description tasks, and an additional language for the Translation task. In addition, we released a new evaluation dataset featuring ambiguities that we expected would benefit from visual context. Table \ref{tab:datasets:overview} presents an overview of the new evaluation datasets. Figure~\ref{fig:examples:en-de-fr} shows an example of an image with an aligned English-German-French description.

In addition to releasing the parallel text, we also distributed two types of ResNet-50 visual features \citep{he2016resnet} for all of the images, namely the `res4\_relu' convolutional features (which preserve the spatial location of a feature in the original image) and averaged pooled features.

\subsection*{Multi30K French Translations}

We extended the translation data in Multi30K dataset with crowdsourced French translations. The crowdsourced translations were collected from 12 workers using an internal platform. We estimate the translation work had a monetary value of \euro 9,700.
The translators had access to the source segment, the image and an automatic translation created with a standard phrase-based system \citep{Koehn2007} trained on WMT'15 parallel text. The automatic translations were presented to the crowdworkers to further simplify the crowdsourcing task. We note that this did not end up being a post-editing task, that is, the translators did not simply copy and paste the suggested translations. To demonstrate this, we calculated text-similarity metric scores between the phrase-based system outputs and the human translations on the training corpus, resulting in 
0.41 edit distance (measured using the TER metric), meaning that more than 40\% of the words between these two versions do not match. 

\subsection*{\testdata}

\begin{table}
\renewcommand{\arraystretch}{1.3}
\centering
\begin{tabular}{lcc}
\toprule
Group & Task 1 & Task 2 \\
\midrule
Strangers! & 150 & 154 \\
Wild Child & 83 & 83\\
Dogs in Action & 78 & 92\\
Action Photography & 238 & 259\\
Flickr Social Club & 241 & 263\\
Everything Outdoor & 206 & 214 \\
Outdoor Activities & 4 & 6 \\
\bottomrule
\end{tabular}
\caption{Distribution of images in the \testdata{} by Flickr group.}\label{tab:data:groups}
\end{table}

We collected new evaluation data for the Multi30K dataset. We sampled new images from five of the six Flickr groups used to create the original Flickr30K dataset using MMFeat \citep{Kiela2016}\footnote{Strangers!, Wild Child, Dogs in Action, Action Photography, and Outdoor Activities.}. We sampled additional images from two thematically related groups (Everything Outdoor and Flickr Social Club) because Outdoor Activities only returned 10 new CC-licensed images and Flickr-Social no longer exists. Table \ref{tab:data:groups} shows the distribution of images across the groups and tasks. We initially downloaded 2,000 images per Flickr group, which were then manually filtered by three of the authors. The filtering was done to remove (near) duplicate images, clearly watermarked images, and images with dubious content. This process resulted in a total of 2,071 images.

We crowdsourced five English descriptions of each image from Crowdflower\footnote{\url{http://www.crowdflower.com}} using the same process as \citet{ElliottFrankSimaanSpecia2016}. One of the authors selected 1,000 images from the collection to form the dataset for the Multimodal Translation task based on a manual inspection of the English descriptions. Professional German translations were collected for those 1,000 English-described images. The remaining 1,071 images were used for the Multilingual Image Description task. We collected five additional independent German descriptions of those images from Crowdflower.

\subsection*{Ambiguous COCO}

\begin{figure*}
\begin{subfigure}[t]{0.47\linewidth}
\begin{center}
\includegraphics[width=0.88\textwidth]{COCO_train2014_000000445192.jpg}
\end{center}
En: A man on a motorcycle is \underline{passing} another vehicle. \\
De: Ein Mann auf einem Motorrad \underline{fährt} an einem anderen Fahrzeug vorbei.
\\
Fr: Un homme sur une moto \underline{dépasse} un autre véhicule.
\end{subfigure}
\hspace{3em}
\begin{subfigure}[t]{0.45\linewidth}
\begin{center}
\includegraphics[width=0.83\textwidth]{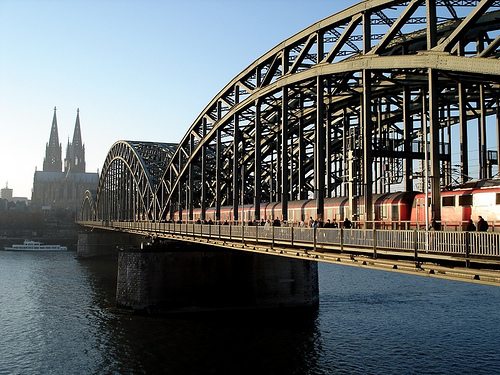}
\end{center}
En: A red train is \underline{passing} over the water on a bridge \\
De: Ein roter Zug \underline{fährt} auf einer Brücke über das Wasser\\
Fr: Un train rouge \underline{traverse} l'eau sur un pont.
\end{subfigure}
   \caption{Two senses of the English verb "to pass" in their visual contexts, with the original English and the translations into German and French. The verb and its translations are underlined.}
    \label{fig:examples_ambiguous_coco}
\end{figure*}

As a secondary evaluation dataset for the Multimodal Translation task, we collected and translated a set of image descriptions that potentially contain ambiguous verbs. We based our selection on the VerSe dataset \citep{gella-lapata-keller:2016:N16-1}, which annotates a subset of the COCO \citep{DBLP:journals/corr/LinMBHPRDZ14} and TUHOI \citep{Le14} images with OntoNotes senses for 90  verbs which are ambiguous, e.g.~{\em play}. Their goals were to test the feasibility of annotating images with the word sense of a given verb (rather than verbs themselves) and to provide a gold-labelled dataset for evaluating automatic visual sense disambiguation methods.

Altogether, the VerSe dataset contains 3,518 images, but we limited ourselves to its COCO section, since for our purposes we also need the image descriptions, which are not available in TUHOI. The COCO portion covers 82 verbs; we further discarded verbs that are unambiguous in the dataset, i.e.~although some verbs have multiple senses in OntoNotes, they all occur with one sense in VerSe (e.g.~\textit{gather} is used in all instances to describe the `people gathering' sense), resulting in 57 ambiguous verbs (2,699 images). The actual descriptions of the images were not distributed with the VerSe dataset. However, given that the  ambiguous verbs were selected based on the image descriptions, we assumed that in all cases at least one of the original COCO description (out of the five per image) should contain the ambiguous verb. In cases where more than one description contained the verb, we randomly selected one such description to be part of the dataset of descriptions containing ambiguous verbs. This resulted in 2,699 descriptions. 

As a consequence of the original goals of the VerSe dataset, each sense of each ambiguous verb was used multiple times in the dataset, which resulted in many descriptions with the same sense, for example, 85 images (and descriptions) were available for the verb {\em show}, but they referred to a small set of senses of the verb.

The number of images (and therefore descriptions) per ambiguous verb varied from 6 ({\em stir}) to 100 ({\em pull}, {\em serve}). Since our intention was to have a small but varied dataset, we selected a subset of a subset of descriptions per ambiguous verb, aiming at keeping 1-3 instances per sense per verb. This resulted in 461 descriptions for 56 verbs in total, ranging from 3 (e.g.~{\em shake}, {\em carry}) to 26 ({\em reach}) (the verb {\em lay/lie} was excluded as it had only one sense).  
We note that the descriptions include the use of the verbs in phrasal verbs. 
Two examples of the English verb ``to pass'' are shown in Figure \ref{fig:examples_ambiguous_coco}. In the German translations, the source language verb did not require disambiguation (both German translations use the verb ``fährt''), whereas in the French translations, the verb was disambiguated into ``dépasse'' and ``traverse'', respectively.

\section{Participants}

This year we attracted submissions from nine different groups. Table \ref{tab:groups-overview} presents an overview of the groups and their submission identifiers.

\begin{table*}[ht!] 
\begin{center}
\renewcommand{\arraystretch}{1.3}
\begin{tabular}{rp{12cm}}
\toprule
ID & Participating team \\
\midrule 
AFRL-OHIOSTATE & Air Force Research Laboratory \& Ohio State University \citep{Gwinnup2017}\\
CMU & Carnegie Melon University \citep{Jaffe2017}\\
CUNI & Univerzita Karlova v Praze \citep{Helcl2017} \\
DCU-ADAPT & Dublin City University \citep{Calixto2017} \\
LIUMCVC &  Laboratoire d'Informatique de l'Universit\'{e} du Maine
\& Universitat Autonoma de Barcelona Computer Vision Center \citep{Caglayan2017WMT}\\
NICT & National Institute of Information and Communications Technology \& Nara Institute of Science and Technology \citep{Zhang2017} \\
OREGONSTATE & Oregon State University \citep{Ba2017} \\
SHEF & University of Sheffield \citep{Madhyastha2017}\\  
UvA-TiCC & Universiteit van Amsterdam \& Tilburg University \citep{ElliottKadar2017}\\  
\bottomrule
\end{tabular}
\end{center}
\caption{Participants in the WMT17 multimodal machine translation shared task.}\label{tab:groups-overview}
\end{table*}

\paragraph{AFRL-OHIOSTATE} (Task 1)
The AFRL-OHIOSTATE system  submission is an atypical Machine Translation (MT) system in that the image is the catalyst for the MT results, and not the textual content.
This system architecture assumes an image caption engine can be trained in a target language to give meaningful output in the form of a set of the most probable $n$ target language candidate captions. A learned mapping function of the encoded source language caption to the corresponding encoded target language captions is then employed. Finally, a distance function is applied to retrieve the ``nearest'' candidate caption to be the translation of the source caption.

\paragraph{CMU} (Task 2)
The CMU submission uses a multi-task learning technique, extending the baseline so that it generates both a German caption and an English caption. First, a German caption is generated using the baseline method. After the LSTM for the baseline model finishes producing a German caption, it has some final hidden state. Decoding is simply resumed starting from that final state with an independent decoder, separate vocabulary, and this time without any direct access to the image. The goal is to encourage the model to keep information about the image in the hidden state throughout the decoding process, hopefully improving the model output. Although the model is trained to produce both German and English captions, at evaluation time the English component of the model is ignored and only German captions are generated.

\paragraph{CUNI} (Tasks 1 and 2)
For Task 1, the submissions employ the standard neural MT (NMT) scheme enriched with another attentive encoder for the input image. It uses a hierarchical attention combination in the decoder \citep{libovicky2017attention}. The best system was trained with additional data obtained from selecting similar sentences from parallel corpora  and by back-translation of similar sentences found in the SDEWAC corpus \citep{faass2013sdewac}.

The submission to Task 2 is a combination of two neural models. The first model generates an English caption from the image. The second model is a text-only NMT model that translates the English caption to German.

\paragraph{DCU-ADAPT} (Task 1) 
This submission evaluates ensembles of up to four different multimodal NMT models. All models use global image features obtained with the pre-trained CNN VGG19, and are either incorporated in the encoder or the decoder. These models are described in detail in \citep{Calixto2017EMNLP}. They are model IMG$_\text{W}$, in which image features are used as words in the source-language encoder; model IMG$_\text{E}$, where image features are used to initialise the hidden states of the forward and backward encoder RNNs; and model IMG$_\text{D}$, where the image features are used as additional signals to initialise the decoder hidden state. Each image has one corresponding feature vector, obtained from the activations of the FC7 layer of the VGG19 network, and consist of a 4096D real-valued vector that encode information about the entire image.

\paragraph{LIUMCVC} (Task 1)
LIUMCVC experiment with two approaches: a multimodal attentive NMT with separate attention \citep{caglayan2016att}
over source text and convolutional image features, and an NMT where global visual features (2048-dimensional pool5 features from ResNet-50) are multiplicatively interacted with word embeddings. 
More specifically, each target word embedding is multiplied with global visual features in an element-wise fashion in order to visually contextualize word representations.
With 128-dimensional embeddings and 256-dimensional recurrent layers, the resulting models have around 5M parameters.

\paragraph{NICT} (Task 1) 
These are constrained submissions for both language pairs. First, a hierarchical phrase-based (HPB) translation system s built using Moses \citep{Koehn2007} with standard features. Then, an attentional encoder-decoder network  \citep{DBLP:journals/corr/BahdanauCB14} is trained and used as an additional feature to rerank the n-best output of the HPB system. A unimodal NMT model is also trained to integrate visual information. Instead of integrating visual features into the NMT model directly, image retrieval methods are employed to obtain target language descriptions of images that are similar to the image described by the source sentence, and this  target description information is integrated into the NMT model. A multimodal NMT model is also used to rerank the HPB output. All feature weights (including the standard features, the NMT feature and the multimodal NMT feature) were tuned by MERT \citep{Och:2003:MERT}. On the development set, the NMT feature improved the HPB system significantly. However, the multimodal NMT feature did not further improve the HPB system that had integrated the NMT feature.

\paragraph{OREGONSTATE} (Task 1)
The OREGONSTATE system uses a very simple but effective model which feeds the image information to both encoder and decoder. On the encoder side, the image representation was used as an initialization information to generate the source words' representations. This step strengthens the relatedness between image's and source words' representations. Additionally,  the decoder uses alignment to source words by a global attention mechanism. In this way, the decoder benefits from both image and source language information and generates more accurate target side sentence. 

\paragraph{UvA-TiCC} (Task 1)
The submitted systems are Imagination models \citep{ElliottKadar2017}, which are trained to perform two tasks in a multitask learning framework: a) produce the target sentence, and b) predict the visual feature vector of the corresponding image. The constrained models are trained over only the 29,000 training examples in the Multi30K dataset with a source-side vocabulary of 10,214 types and a target-side vocabulary of 16,022 types. The unconstrained models are trained over a concatenation of the Multi30K, News Commentary \citep{Tiedemann2012} parallel texts, and MS COCO \citep{Chen2015} dataset with a joint source-target vocabulary of 17,597 word pieces \citep{SchusterNakajima2012}. In both constrained and unconstrained submissions, the models were trained to predict the 2048D GoogleLeNetV3 feature vector  \citep{Szegedy2015} of an image associated with a source language sentence. The output of an ensemble of the three best randomly initialized models - as measured by BLEU on the Multi30K development set - was used for both the constrained and unconstrained submissions.

\paragraph{SHEF} (Task 1)
The SHEF systems utilize the predicted posterior probability distribution over the image object
classes as image features. To do so, they make use of the  pre-trained ResNet-152~\citep{he2016resnet}, a deep CNN based image network that is trained over the 1,000 object categories on the Imagenet dataset~\citep{imagenet_cvpr09} to obtain the posterior distribution.
The model follows a standard encoder-decoder NMT approach using \emph{softdot} attention as
described in~\citep{luong2015effective}. It explores image
information in three ways: a) to initialize the encoder; b) to initialize the decoder; c) to condition each source word with the image class posteriors. In all these three  ways, non-linear affine transformations over the posteriors are used as image features.

\paragraph{Baseline --- Task 1}
The baseline system for the multimodal translation task is a text-only neural machine translation system built with the  Nematus toolkit \citep{Sennrich2017}. Most settings and hyperparameters were kept as default, with a few exceptions: batch size of 40 (instead of 80 due to memory constraints) and ADAM as optimizer. In order to handle rare and OOV words, we used the Byte Pair Encoding Compression Algorithm to segment words \citep{Sennrich:2016}. The merge operations for word segmentation were learned using training data in both source and target languages. These were then applied to all training, validation and test sets in both source and target languages. In post-processing, the original words were restored by concatenating the subwords.

\paragraph{Baseline --- Task 2}
The baseline for the multilingual image description task is an attention-based image description system trained over only the German image descriptions \citep{Caglayan2017}.
The visual representation are extracted from the so-called \emph{res4f\_relu} layer from a ResNet-50 \citep{he2016resnet} convolutional neural network trained on the ImageNet dataset \citep{ILSVRC15}.
Those feature maps provide spatial information on which the model focuses through the attention mechanism.

\section{Text-similarity Metric Results}\label{sec:results}

The submissions were evaluated against either professional or crowd-sourced references. All submissions and references were pre-processed to lowercase, normalise punctuation, and tokenise the sentences using the Moses scripts.\footnote{\url{https://github.com/moses-smt/mosesdecoder/blob/master/scripts/}} The evaluation was performed using {\tt MultEval} \citep{Clark2011} with the primary metric of Meteor 1.5 \citep{Denkowski2014}. We also report the results using BLEU \citep{Papineni:2002:BMA:1073083.1073135} and TER \citep{snover2006study} metrics. The winning submissions are indicated by $\bullet$. These are the top-scoring submissions and those that are not significantly different (based on Meteor scores) according the approximate randomisation test (with p-value $\leq$ 0.05) provided by {\tt MultEval}. Submissions marked with * are not significantly different from the Baseline according to the same test.

\subsection{Task 1: English $\rightarrow$ German}

\subsubsection{\testdata}

Table \ref{tab:t1-ende-flickr-results} shows the results on the \testdata{} with a German target language. It interesting to note that the metrics do not fully agree on the ranking of systems, although the four best (statistically indistinguishable) systems win by all metrics. All-but-one submission outperformed the text-only NMT baseline. This year, the best performing systems include both multimodal (LIUMCVC\_MNMT\_C and UvA-TiCC\_IMAGINATION\_U) and text-only (NICT\_NMTrerank\_C and LIUMCVC\_MNMT\_C) submissions. (Strictly speaking, the UvA-TiCC\_IMAGINATION\_U system is incomparable because it is an unconstrained system, but all unconstrained systems perform in the same range as the constrained systems.)

\begin{table*}
  \centering
  \renewcommand{\arraystretch}{1.2}
  \begin{tabular}{lccc}
  \toprule
  & BLEU $\uparrow$ & Meteor $\uparrow$ & TER $\downarrow$\\
  \midrule
  $\bullet$LIUMCVC\_MNMT\_C & 33.4 & 54.0 & 48.5 \\
$\bullet$NICT\_NMTrerank\_C & 31.9 & 53.9 & 48.1 \\
$\bullet$LIUMCVC\_NMT\_C & 33.2 & 53.8 & 48.2 \\
\rowcolor[gray] {.9} $\bullet$UvA-TiCC\_IMAGINATION\_U & 33.3 & 53.5 & 47.5 \\
UvA-TiCC\_IMAGINATION\_C & 30.2 & 51.2 & 50.8 \\
\rowcolor[gray] {.9} CUNI\_NeuralMonkeyTextualMT\_U & 31.1 & 51.0 & 50.7 \\
OREGONSTATE\_2NeuralTranslation\_C & 31.0 & 50.6 & 50.7 \\
DCU-ADAPT\_MultiMT\_C & 29.8 & 50.5 & 52.3 \\
\rowcolor[gray] {.9} CUNI\_NeuralMonkeyMultimodalMT\_U & 29.5 & 50.2 & 52.5 \\
CUNI\_NeuralMonkeyTextualMT\_C & 28.5 & 49.2 & 54.3 \\
OREGONSTATE\_1NeuralTranslation\_C & 29.7 & 48.9 & 51.6 \\
CUNI\_NeuralMonkeyMultimodalMT\_C & 25.8 & 47.1 & 56.3 \\
SHEF\_ShefClassInitDec\_C & 25.0 & 44.5 & 53.8 \\
SHEF\_ShefClassProj\_C & 24.2 & 43.4 & 55.9 \\
Baseline (text-only NMT)& 19.3 & 41.9 & 72.2 \\
\rowcolor[gray] {.9} AFRL-OHIOSTATE-MULTIMODAL\_U & 6.5 & 20.2 & 87.4 \\
  \bottomrule
  \end{tabular}
  \caption{\label{tab:t1-ende-flickr-results} Official results for the WMT17 Multimodal Machine Translation task on the English-German \testdata. Systems with grey background indicate use of resources that fall outside the constraints provided for the shared task.}

\end{table*}

\subsubsection{Ambiguous COCO}

Table \ref{tab:t1-ende-coco-results} shows the results for the out-of-domain ambiguous COCO dataset with a German target language. Once again the evaluation metrics do not fully agree on the ranking of the submissions. 

It is interesting to note that the metric scores are lower for the out-of-domain Ambiguous COCO data compared to the in-domain \testdata.  However, we cannot make definitive claims about the difficulty of the dataset because the Ambiguous COCO dataset contains fewer sentences than the \testdata{} (461 compared to 1,000).

The systems are mostly in the same order as on the \testdata, with the same four systems performing best. However, two systems (DCU-ADAPT\_MultiMT\_C and OREGONSTATE\_1NeuralTranslation\_C) are ranked higher on this test set than on the in-domain Flickr dataset, indicating that they are relatively more robust and possibly better at resolving the ambiguities found in the Ambiguous COCO dataset.

\begin{table*}
  \centering
  \renewcommand{\arraystretch}{1.2}
  \begin{tabular}{lccc}
  \toprule
  & BLEU $\uparrow$ & Meteor $\uparrow$ & TER $\downarrow$\\
  \midrule
$\bullet$LIUMCVC\_NMT\_C & 28.7 & 48.9 & 52.5 \\
$\bullet$LIUMCVC\_MNMT\_C & 28.5 & 48.8 & 53.4 \\
$\bullet$NICT\_1\_NMTrerank\_C & 28.1 & 48.5 & 52.9 \\
\rowcolor[gray] {.9} $\bullet$UvA-TiCC\_IMAGINATION\_U & 28.0 & 48.1 & 52.4 \\
DCU-ADAPT\_MultiMT\_C & 26.4 & 46.8 & 54.5 \\
OREGONSTATE\_1NeuralTranslation\_C & 27.4 & 46.5 & 52.3 \\
\rowcolor[gray] {.9} CUNI\_NeuralMonkeyTextualMT\_U & 26.6 & 46.0 & 54.8 \\
UvA-TiCC\_IMAGINATION\_C & 26.4 & 45.8 & 55.4 \\
OREGONSTATE\_2NeuralTranslation\_C & 26.1 & 45.7 & 55.9 \\
\rowcolor[gray] {.9} CUNI\_NeuralMonkeyMultimodalMT\_U & 25.7 & 45.6 & 55.7 \\
CUNI\_NeuralMonkeyTextualMT\_C & 23.2 & 43.8 & 59.8 \\
CUNI\_NeuralMonkeyMultimodalMT\_C & 22.4 & 42.7 & 60.1 \\
SHEF\_ShefClassInitDec\_C & 21.4 & 40.7 & 56.5 \\
SHEF\_ShefClassProj\_C & 21.0 & 40.0 & 57.8 \\
Baseline (text-only NMT)& 18.7 & 37.6 & 66.1 \\
  \bottomrule
  \end{tabular}
  \caption{\label{tab:t1-ende-coco-results} Results for the Multimodal Translation task on the English-German Ambiguous COCO dataset. 
Systems with grey background indicate use of resources that fall outside the constraints provided for the shared task.}

\end{table*}

\subsection{Task 1: English $\rightarrow$ French}

\subsubsection{Multi30K Test 2017}

Table \ref{tab:t1-enfr-flickr-results} shows the results for the \testdata{} with French as target language. A reduced number of submissions were received for this new language pair, with no unconstrained systems. In contrast to the English$\rightarrow$German results, the evaluation metrics are in better agreement about the ranking of the submissions. 

Translating from English$\rightarrow$French is an easier task than English$\rightarrow$German systems, as reflected in the higher metric scores.
This also includes the baseline systems where English$\rightarrow$French results in 63.1 Meteor compared to 41.9 for English$\rightarrow$German. 

Eight out of the ten submissions outperformed the English$\rightarrow$French baseline system. Two of the best submissions for English$\rightarrow$German remain the best for English$\rightarrow$French (LIUMCVC\_MNMT\_C and NICT\_NMTrerank\_C), the text-only system (LIUMCVC\_NMT\_C) decreased in performance, and no UvA-TiCC\_IMAGINATION\_U system was submitted for French.

An interesting observation is the difference of the Meteor scores between text-only NMT system (LIUMCVC\_NMT\_C) and Moses  
hierarchical phrase-based system with reranking (NICT\_NMTrerank\_C). While the two systems are very close for the English$\rightarrow$German direction, the hierarchical system is better than the text-only NMT systems in the English$\rightarrow$French direction. This pattern holds for both the \testdata{} and Ambiguous COCO test data.
\begin{table*}
  \centering
  \renewcommand{\arraystretch}{1.2}
  \begin{tabular}{lccc}
  \toprule
  & BLEU $\uparrow$ & Meteor $\uparrow$ & TER $\downarrow$\\
  \midrule
$\bullet$LIUMCVC\_MNMT\_C & 55.9 & 72.1 & 28.4 \\
$\bullet$NICT\_NMTrerank\_C & 55.3 & 72.0 & 28.4 \\
DCU-ADAPT\_MultiMT\_C & 54.1 & 70.1 & 30.0 \\
LIUMCVC\_NMT\_C & 53.3 & 70.1 & 31.7 \\
OREGONSTATE\_2NeuralTranslation\_C & 51.9 & 68.3 & 32.7 \\
OREGONSTATE\_1NeuralTranslation\_C & 51.0 & 67.2 & 33.6 \\
CUNI\_NeuralMonkeyMultimodalMT\_C & 49.9 & 67.2 & 34.3 \\
CUNI\_NeuralMonkeyTextualMT\_C & 50.3 & 67.0 & 33.6 \\
Baseline (text-only NMT) & 44.3 & 63.1 & 39.6 \\
*SHEF\_ShefClassInitDec\_C & 45.0 & 62.8 & 38.4 \\
SHEF\_ShefClassProj\_C & 43.6 & 61.5 & 40.5 \\
  \bottomrule
  \end{tabular}
  \caption{\label{tab:t1-enfr-flickr-results} Results for the Multimodal Translation task on the English-French Multi30K Test 2017 data.}

\end{table*}

\subsubsection{Ambiguous COCO}

Table \ref{tab:t1-enfr-coco-results} shows the results for the out-of-domain Ambiguous COCO dataset with the French target language. Once again, in contrast to the English$\rightarrow$German results, the evaluation metrics are in better agreement about the ranking of the submissions. The performance of all the models is once again in mostly agreement with the \testdata, albeit lower. Both DCU-ADAPT\_MultiMT\_C and OREGONSTATE\_2NeuralTranslation\_C again perform relatively better on this dataset.

\begin{table*}
  \centering
  \renewcommand{\arraystretch}{1.2}
  \begin{tabular}{lccc}
  \toprule
  & BLEU $\uparrow$ & Meteor $\uparrow$ & TER $\downarrow$\\
  \midrule
$\bullet$LIUMCVC\_MNMT\_C & 45.9 & 65.9 & 34.2 \\
$\bullet$NICT\_NMTrerank\_C & 45.1 & 65.6 & 34.7 \\
$\bullet$DCU-ADAPT\_MultiMT\_C & 44.5 & 64.1 & 35.2 \\
OREGONSTATE\_2NeuralTranslation\_C & 44.1 & 63.8 & 36.7 \\
LIUMCVC\_NMT\_C & 43.6 & 63.4 & 37.4 \\
CUNI\_NeuralMonkeyTexutalMT\_C & 43.0 & 62.5 & 38.2 \\
CUNI\_NeuralMonkeyMultimodalMT\_C & 42.9 & 62.5 & 38.2 \\
OREGONSTATE\_1NeuralTranslation\_C & 41.2 & 61.6 & 37.8 \\
SHEF\_ShefClassInitDec\_C & 37.2 & 57.3 & 42.4 \\
*SHEF\_ShefClassProj\_C & 36.8 & 57.0 & 44.5 \\
Baseline (text-only NMT) & 35.1 & 55.8 & 45.8 \\
  \bottomrule
  \end{tabular}
  \caption{\label{tab:t1-enfr-coco-results} Results for the Multimodal Translation task on the English-French Ambiguous COCO dataset.}
\end{table*}

\subsection{Task 2: English $\rightarrow$ German}

\begin{table*}[t]
  \centering
  \renewcommand{\arraystretch}{1.2}
  \begin{tabular}{lccc}
  \toprule
  & BLEU $\uparrow$ & Meteor $\uparrow$ & TER $\downarrow$\\
  \midrule
Baseline (target monolingual) & 9.1 & 23.4 & 91.4 \\
CUNI\_NeuralMonkeyCaptionAndMT\_C & 4.2 & 22.1 & 133.6 \\
CUNI\_NeuralMonkeyCaptionAndMT\_U & 6.5 & 20.6 & 91.7 \\
CMU\_NeuralEncoderDecoder\_C & 9.1 & 19.8 & 63.3 \\
CUNI\_NeuralMonkeyBilingual\_C & 2.3 & 17.6 & 112.6 \\
  \bottomrule
  \end{tabular}
  \caption{\label{t2:results} Results for the Multilingual Image Description task on the English-German \testdata.}
\end{table*}

The description generation task, in which systems must generate target-language (German) captions for a test image, has substantially changed since last year. The main difference is that source language descriptions are no longer observed for images at test time. The training data remains the same and contains images with both source and target language descriptions. The aim is thus to leverage multilingual training data to improve a monolingual task.

Table \ref{t2:results} shows the results for the Multilingual image description task. This task attracted fewer submissions than last year, which may be because it was no longer possible to re-use a model designed for Multimodal Translation. The evaluation metrics do not agree on the ranking of the submissions, with major differences in the ranking using either BLEU or TER instead of Meteor. 

The main result is that none of the submissions outperform the monolingual German baseline according to Meteor. All of the submissions are statistically significantly different compared to the  baseline. However, the CMU\_NeuralEncoderDecoder\_C submission marginally outperformed the baseline according to TER and equalled its BLEU score.

\section{Human Judgement Results}\label{sec:human}

This year, we conducted a human evaluation in addition to the text-similarity metrics to assess the translation quality of the submissions. This evaluation was undertaken for the Task 1 German and French outputs for the \testdata.

This section describes how we collected the human assessments and computed the results. We would like to gratefully thank all assessors. 

\begin{figure*}
\centering
\includegraphics[width=0.9\textwidth]{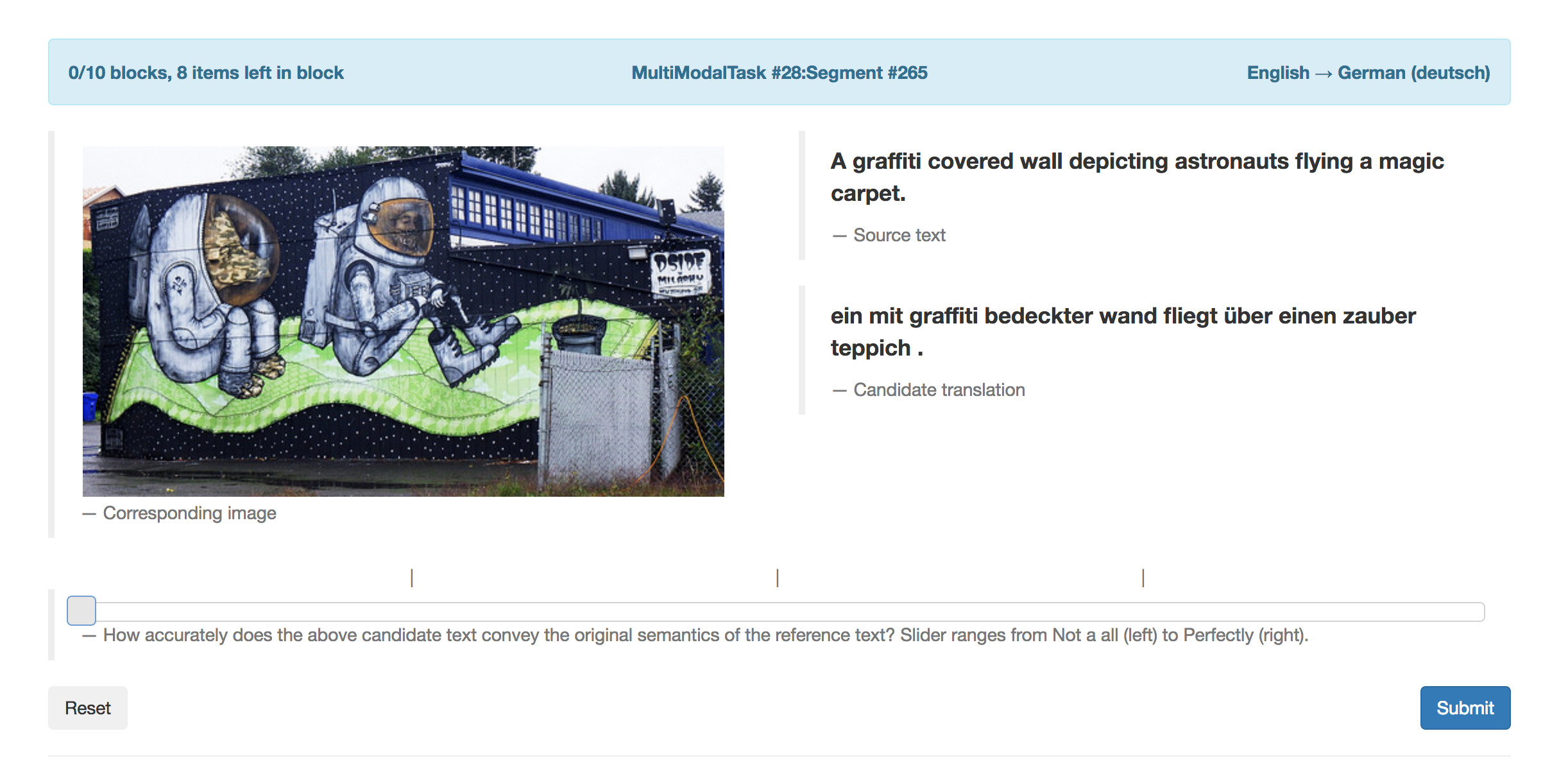}
\caption{Example of the human direct assessment evaluation interface.}\label{fig:human:example}
\end{figure*}

\subsection{Methodology}
The system outputs were manually evaluated by bilingual Direct Assessment (DA) \citep{graham2017can} using the Appraise platform \citep{appraise}.
The annotators (mostly researchers) were asked to evaluate the semantic relatedness between the source sentence in English and the target sentence in  German or French. 
The image was shown along with the source sentence and the candidate translation and evaluators were told to rely on the image when necessary to obtain a better understanding of the source sentence (e.g. in cases where the text was ambiguous).
Note that the reference sentence is not displayed during the evaluation, in order to avoid influencing the assessor. Figure \ref{fig:human:example} shows an example of the direct assessment interface used in the evaluation.
The score of each translation candidate ranges from 0 (meaning that the meaning of the source is not preserved in the target language sentence) to 100 (meaning the meaning of the source is ``perfectly'' preserved).
The human assessment scores are standardized according to each individual assessor's overall mean and standard deviation score. 
The overall score of a given system ($z$) corresponds to the mean standardized score of its translations.

\subsection{Results}

The French outputs were evaluated by seven assessors, who conducted a total of 2,521 DAs, resulting in a minimum of 319 and a maximum of 368 direct assessments per system submission, respectively.
The German outputs were evaluated by 25 assessors, who conducted a total of 3,485 DAs, resulting in a minimum of 291 and a maximum of 357 direct assessments per system submission, respectively. This is somewhat less than the recommended number of 500, so the results should be considered preliminary.

Tables~\ref{tab:t1-ende-flickr-human} and~\ref{tab:t1-enfr-flickr-human} show the results of the human evaluation for the English to German and the English to French Multimodal Translation task (\testdata). 
The systems are ordered by standardized mean DA scores and clustered according to the Wilcoxon signed-rank test at p-level p $\leq$ 0.05. Systems within a cluster are considered tied. The Wilcoxon signed-rank scores can be found in Tables~\ref{tab:ende_wilcoxon} and \ref{tab:enfr_wilcoxon} in Appendix~A.

\definecolor{text}{HTML}{00008B}
\definecolor{multi}{HTML}{007777}

\begin{table*}[htb]
\centering
\begin{tabular}{llll}
\multicolumn{4}{c}{\textbf{English}$\rightarrow$\textbf{German}} \\
\# & Raw & $z$ & System  \\
\toprule
1 & 77.8 & 0.665 & LIUMCVC\_MNMT\_C \\
 \midrule
\rowcolor[gray]{0.9} 2 & 74.1 & 0.552 & UvA-TiCC\_IMAGINATION\_U \\
\midrule
3 & 70.3 & 0.437 & NICT\_NMTrerank\_C \\ 
\rowcolor[gray]{0.9} & 68.1 & 0.325 & CUNI\_NeuralMonkeyTextualMT\_U \\
& 68.1 & 0.311 & DCU-ADAPT\_MultiMT\_C \\
& 65.1 & 0.196 & LIUMCVC\_NMT\_C \\
\rowcolor[gray]{0.9} & 60.6 & 0.136 & CUNI\_NeuralMonkeyMultimodalMT\_U \\
& 59.7 & 0.08 & UvA-TiCC\_IMAGINATION\_C \\
& 55.9 & -0.049 & CUNI\_NeuralMonkeyMultimodalMT\_C \\
& 54.4 & -0.091 & OREGONSTATE\_2NeuralTranslation\_C \\
& 54.2 & -0.108 & CUNI\_NeuralMonkeyTextualMT\_C \\
& 53.3 & -0.144 & OREGONSTATE\_1NeuralTranslation\_C \\
& 49.4 & -0.266 & SHEF\_ShefClassProj\_C \\
& 46.6 & -0.37 & SHEF\_ShefClassInitDec\_C \\
 \midrule
15 & 39.0 & -0.615 & Baseline (text-only NMT)\\
\rowcolor[gray]{0.9} & 36.6 & -0.674 & AFRL-OHIOSTATE\_MULTIMODAL\_U \\
\bottomrule
\end{tabular}
\caption{\label{tab:t1-ende-flickr-human} Results of the human evaluation of the WMT17 English-German Multimodal Translation task (\testdata). Systems are ordered by standardized mean DA scores ($z$) and clustered according to Wilcoxon signed-rank test at p-level p $\leq$ 0.05. Systems within a cluster are considered tied, although systems within a cluster may be statistically significantly different from each other (see Table~\ref{tab:ende_wilcoxon}).
Systems using unconstrained data are identified with a gray background.
}
\end{table*}

\begin{table*}[htb]
\centering
\begin{tabular}{llll}
\multicolumn{4}{c}{\textbf{English}$\rightarrow$\textbf{French}} \\
\# & Raw & $z$ & System  \\
\toprule
1 & 79.4 & 0.446 & NICT\_NMTrerank\_C \\
& 74.2 & 0.307 & CUNI\_NeuralMonkeyMultimodalMT\_C \\
& 74.1 & 0.3 & DCU-ADAPT\_MultiMT\_C \\
 \midrule
4 & 71.2 & 0.22 & LIUMCVC\_MNMT\_C \\
& 65.4 & 0.056 & OREGONSTATE\_2NeuralTranslation\_C \\
& 61.9 & -0.041 & CUNI\_NeuralMonkeyTextualMT\_C \\
& 60.8 & -0.078 & OREGONSTATE\_1NeuralTranslation\_C \\
& 60.5 & -0.079 & LIUMCVC\_NMT\_C \\
 \midrule
9 & 54.7 & -0.254 & SHEF\_ShefClassInitDec\_C \\
& 54.0 & -0.282 & SHEF\_ShefClassProj\_C \\
 \midrule
11 & 44.1 & -0.539 & Baseline (text-only NMT)\\
\bottomrule
\end{tabular}
\caption{\label{tab:t1-enfr-flickr-human} Results of the human evaluation of the WMT17 English-French Multimodal Translation task (\testdata). Systems are ordered by standardized mean DA score ($z$) and clustered according to Wilcoxon signed-rank test at p-level p ≤ 0.05. Systems within a cluster are considered tied, although systems within a cluster may be statistically significantly different from each other (see Table~\ref{tab:enfr_wilcoxon}).}
\end{table*}

\begin{figure*}
\centering
  \begin{tikzpicture}
    \begin{axis}[xlabel={Human judgments ($z$)}, ylabel=Meteor,
      axis x line*=bottom,
      axis y line*=left,
      xmin = -1, xmax=1,
      ymin = 40, ymax=55,
      width=\textwidth*0.8,
      height=\textwidth*0.56,
      ]
      \addplot[scatter, only marks, 
      mark=*, black,
      point meta= explicit symbolic,
      nodes near coords, 
      label style={fontsize=\tiny, font=\sansmath\sffamily},
      font=\scriptsize,
      visualization depends on=\thisrow{alignment} \as \alignment,
      every node near coord/.style={anchor=\alignment},
      ]
      table[x=z,y=meteor, meta index=2] {
                        z       meteor      System  class   alignment
   0.665     54.0     LIUMCVC\_MNMT\_C                  0     180
   0.437     53.9     NICT\_NMTrerank\_C                0     -90
   0.196     53.8     LIUMCVC\_NMT\_C                   0     0
   0.552     53.5     UvA-TiCC\_IMAGINATION\_U          0     90
   0.08     51.2      UvA-TiCC\_IMAGINATION\_C          0     -90
   0.325     51.0     CUNI\_NeuralMonkeyTextualMT\_U    0     180
   -0.091     50.6     OREGONSTATE\_2NeuralTranslation\_C 0   0
   0.311     50.5     DCU-ADAPT\_MultiMT\_C             0     180
   0.136     50.2     CUNI\_NeuralMonkeyMultimodalMT\_U 0     175
   -0.108    49.2     CUNI\_NeuralMonkeyTextualMT\_C    0     180
   -0.144     48.9     OREGONSTATE\_1NeuralTranslation\_C 0   120
   -0.049     47.1     CUNI\_NeuralMonkeyMultimodalMT\_C 0    180
   -0.37     44.5     SHEF\_ShefClassInitDec\_C          0    180
   -0.266     43.4     SHEF\_ShefClassProj\_C            0    180
   -0.615     41.9     Baseline           0         180
   };
   \end{axis}
   \end{tikzpicture}
\caption{System performance on the English$\rightarrow$German \testdata{} as measured by human evaluation against Meteor scores. The AFRL-OHIOSTATE-MULTIMODAL\_U system has been ommitted for readability.}\label{da-meteor-corr-de}
\end{figure*}
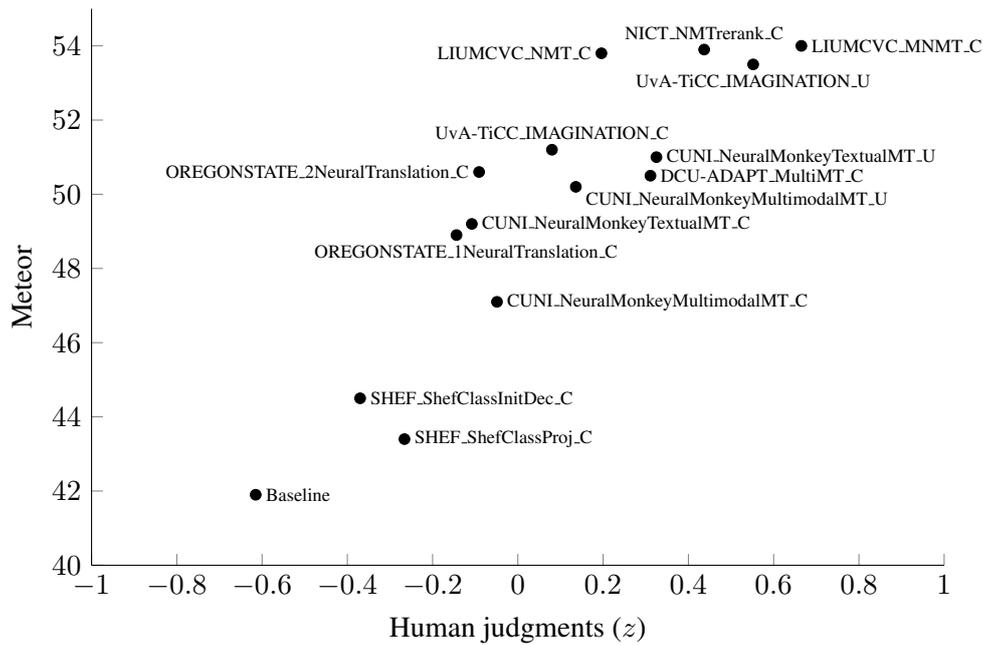

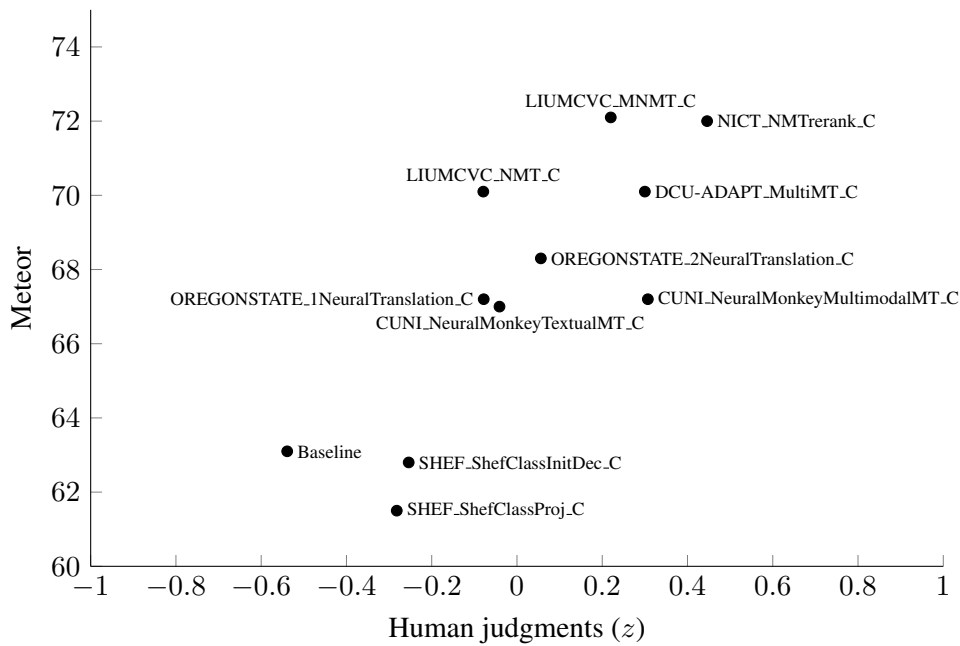
\begin{figure*}
  \centering
    \begin{tikzpicture}
      \begin{axis}[xlabel={Human judgments ($z$)}, ylabel=Meteor,
        axis x line*=bottom,
        axis y line*=left,
        xmin = -1, xmax=1,
        ymin = 60, ymax=75,
        width=\textwidth*0.8,
        height=\textwidth*0.56,
        ]
        \addplot[scatter, only marks, 
        mark=*, black,
        point meta= explicit symbolic,
        nodes near coords, 
        label style={fontsize=\tiny, font=\sansmath\sffamily},
        font=\scriptsize,
        visualization depends on=\thisrow{alignment} \as \alignment,
        every node near coord/.style={anchor=\alignment},
        ]
 table[x=z,y=meteor, meta index=2] {
 z       meteor      System  class   alignment
 0.446     72.0      NICT\_NMTrerank\_C                 0 180
0.307     67.2      CUNI\_NeuralMonkeyMultimodalMT\_C  0 180
0.3       70.1      DCU-ADAPT\_MultiMT\_C              0 180
0.22      72.1      LIUMCVC\_MNMT\_C                   0 -90
0.056     68.3      OREGONSTATE\_2NeuralTranslation\_C 0 180
-0.041    67.0      CUNI\_NeuralMonkeyTextualMT\_C     0 120
-0.078    67.2      OREGONSTATE\_1NeuralTranslation\_C 0 0
-0.079    70.1      LIUMCVC\_NMT\_C                    0 -90
-0.254    62.8      SHEF\_ShefClassInitDec\_C          0 180
-0.282    61.5      SHEF\_ShefClassProj\_C             0 180
-0.539    63.1      Baseline                           0 180
     };
     \end{axis}
     \end{tikzpicture}
  \caption{System performance on the
    English$\rightarrow$French \testdata{} as measured by human
  evaluation against Meteor scores.}\label{da-meteor-corr-fr}
 \end{figure*}

When comparing automatic and human evaluations, we can observe that they globally agree with each other, as shown in Figures~\ref{da-meteor-corr-de} and \ref{da-meteor-corr-fr}, with German showing better agreement than French.
We point out two interesting disagreements:
First, in the English$\rightarrow$French language pair, 
CUNI\_NeuralMonkeyMultimodalMT\_C and DCU-ADAPT\_MultiMT\_C are significantly better than LIUMCVC\_MNMT\_C, 
despite the fact that the latter system achieves much higher metric scores.
Secondly, across both languages, the text-only LIUMCVC\_NMT\_C system performs well on metrics but does relatively poorly on human judgements, especially as compared to the multimodal version of the same system. 

\section{Discussion}

\paragraph{Visual Features: do they help?}

Three teams provided text-only counterparts to their multimodal systems for Task 1
(CUNI, LIUMCVC, and OREGONSTATE), which enables us to evaluate the contribution of
visual features. For many systems, visual features did not seem to
help reliably, at least as measured by metric evaluations: in German,
the CUNI and OREGONSTATE text-only systems outperformed the counterparts,
while in French, there were small improvements for the CUNI multimodal
system. However, the LIUMCVC multimodal system outperformed their text-only system across both languages. 

The human evaluation results are perhaps more promising: nearly all
the highest ranked systems (with the exception of NICT) are 
multimodal. An intruiging result was the text-only LIUMCVC\_NMT\_C, which ranked highly on metrics but poorly in the human evaluation. The LIUMCVC systems were indistinguishable from each other in terms of Meteor scores but the standardized mean direct assessment score showed a significant difference in performance (see Tables~\ref{tab:ende_wilcoxon} and~\ref{tab:enfr_wilcoxon}): further analysis of the reasons for humans disliking the text-only translations will be necessary.

The multimodal Task 1 submissions can be broadly categorised into three groups based on how they use the images: approaches useing double-attention mechanisms, initialising the hidden state of the encoder and/or decoder networks with the global image feature vector, and alternative uses of image features. 
The double-attention models calculate context vectors over the source language hidden states and location-preserving feature vectors over the image; these vectors are used as inputs to the translation decoder (CUNI\_NeuralMonkeyMultimodalMT). Encoder and/or decoder initialisation involves initialising the recurrent neural network with an affine transformation of a global image feature vector (DCU-ADAPT\_MultiMT, OREGONSTATE\_1NeuralTranslation) or initialising the encoder and decoder with the 1000 dimension softmax probability vector over the object classes in ImageNet object recognition challenge (SHEF\_ShefClassInitDec). The alternative uses of the image features include element-wise multiplication of the target language embeddings with an affine transformation of a global image feature vector (LIUMCVC\_MNMT), summing the source language word embeddings with affine-transformed 1000 dimension softmax probability vector (SHEF\_ShefClassProj), using the visual features in a retrieval framework (AFRL-OHIOSTATE\_MULTIMODAL), and learning visually-grounded encoder representations by learning to predict the global image feature vector from the source language hidden states (UvA-TiCC\_IMAGINATION).

Overall, the metric and human judgement results in Sections \ref{sec:results} and \ref{sec:human} indicate that there is still a wide scope for exploration of the best way to integrate visual and textual information. In particular, the alternative approaches proposed in the LIUMCVC\_MNMT and UvA-TiCC\_IMAGINATION submissions led to strong performance in both the metric and human judgement results, surpassing the more common approaches using initialisation and double attention.

Finally, the text-only NICT system ranks highly across both languages.
This system uses hierarchical phrase-based MT with a reranking step based on a
neural text-only system, since their multimodal system never
outperformed the text-only variant in development~\citep{Zhang2017}.        
This is in line with last year's results and the strong Moses baseline
\citep{Specia2016}, and suggests a continuing role for phrase-based
MT for small homogeneous datasets. 

\paragraph{Unconstrained systems} 
The Multi30k dataset is relatively small, so unconstrained systems 
use more data to complement the image description translations.
Three groups submitted systems using external resources: UvA-TiCC, CUNI, and AFRL-OHIOSTATE. The unconstrained UvA-TiCC and CUNI submissions
always outperformed their respective constrained variants by 2--3
Meteor points and achieved higher standardized mean DA scores. These results suggest that external parallel text
corpora (UvA-TiCC and CUNI) and external monolingual image description
datasets (UvA-TiCC) can usefully improve the quality of multimodal
translation models.

However, tuning to the target domain remains important, even for relatively simple image captions. We ran the best-performing English$\rightarrow$German WMT'16 news translation system \citep{Sennrich2016WMT} on the English$\rightarrow$German \testdata{} to gauge the performance of a state-of-the-art text-only translation system trained on only out-of-domain resources\footnote{\url{http://data.statmt.org/rsennrich/wmt16_systems/en-de/}}.
It ranked 10th in terms of Meteor (49.9) and 11th in terms of BLEU (29.0), placing it firmly in the middle of the pack, and below nearly all the text-only submissions trained on the in-domain Multi30K dataset.

\paragraph{The effect of OOV words} 
The Multi30k translation training and test data are very similar, 
with a low OOV rate in the Flickr test set (1.7\%). 
In the 2017 test set, 16\% of English test sentences
include a OOV word. 
Human evaluation gave the impression that these often led to errors propagated
throughout the whole sentence.
Unconstrained systems may perform better by having larger vocabularies, as well as more robust statistics.
When we evaluate the English$\rightarrow$German systems over only the 161 OOV-containing test sentences, the highest
ranked submission by all metrics is the unconstrained UvA-TiCC\_IMAGINATION submission,
with +2.5 Metor and +2.2 BLEU over the second best system (LIUMCVC\_NMT; \@45.6 vs~43.1 Meteor and 24.0 vs~21.8 BLEU).

The difference over non-OOV-containing sentences is not nearly as
stark, with constrained systems all performing best (both LIUMCVC
systems, MNMT and NMT, with 56.6 and 56.3 Meteor, respectively)
but unconstrained systems following close behind (UvA-TiCC with 55.4
Meteor, CUNI with 53.4 Meteor).

\paragraph{Ambiguous COCO dataset}

We introduced a new evaluation dataset this year with the aim of
testing systems' ability to use visual features to identify word
senses. 

However, it is unclear whether visual features improve performance on this test set. The text-only NICT\_NMTrerank system
performs competitively, ranking in the top three submissions for both languages.
We find mixed results for submissions with text-only and multimodal counterparts (CUNI, LIUMCVC, OREGONSTATE): LIUMCVC's multimodal
system improves over the text-only system for French but not German,
while the visual features help for German but not French in the CUNI
and OREGONSTATE systems.

We plan to perform a further analysis on the extent of translation ambiguity in this dataset.
We will also continue to work on other methods
for constructing datasets in which textual ambiguity can be disambiguated
by visual information.

\paragraph{Multilingual Image Description}

It proved difficult for Task 2 systems to use the
English data to improve over the monolingual German baseline.

In future iterations of the task, we will consider a lopsided data
setting, in which there is much more English data than target language
data. This setting is more realistic and will push the use of 
multilingual data. We also hope to conduct human evaluation to better 
assess performance because automatic metrics are problematic for 
this task \citep{elliott2014comparing,kilickaya2017re}.

\section{Conclusions}

We presented the results of the second shared task on multimodal translation and multilingual image description. The shared task attracted submissions from nine groups, who submitted a total of 19 systems across the tasks. The Multimodal Translation task attracted the majority of the submissions. Human judgements for the translation task were collected for the first time this year and ranked systems broadly in line with the automatic metrics.

The main findings of the shared task are: 
\begin{enumerate}[label=(\roman*)]
\item There is still scope for novel approaches to integrating visual and linguistic features in multilingual multimodal models, as demonstrated by the winning systems.

\item External resources have an important role to play in improving the performance of multimodal translation models beyond what can be learned from limited training data.

\item The differences between text-only and multimodal systems are being obfuscated by the well-known shortcomings of text-similarity metrics. Multimodal systems often seem to be prefered by humans but not rewarded by metrics.
Future research on this topic, encompassing both multimodal translation and multilingual image description, should be evaluated using human judgements.
\end{enumerate}

In future editions of the task, we will encourage participants to submit the output of single decoder systems to better understand the empirical differences between approaches. 
We are also considering a Multilingual Multimodal Translation challenge, where the systems can observe two language inputs alongside the image to encourage the development of multi-source multimodal models.
\section*{Acknowledgements}

This work was supported by the CHIST-ERA M2CR project
(French National Research Agency No. ANR-15-CHR2-0006-01 -- Loïc Barrault and Fethi Bougares), and 
by the MultiMT project (EU H2020 ERC Starting Grant No. 678017 -- Lucia Specia).
Desmond Elliott acknowledges the support of NWO Vici Grant No. 277-89-002 awarded to K. Sima'an, and an Amazon Academic Research Award.
We thank Josiah Wang for his help in selecting the Ambiguous COCO dataset.

\section*{A Significance tests}

Tables~\ref{tab:ende_wilcoxon} and \ref{tab:enfr_wilcoxon} show the Wilcoxon signed-rank test used to create the clustering of the systems.

\begin{sidewaystable*}
\centering
\scriptsize
\begin{tabular}{lllllllllllllllll}
\toprule
\multicolumn{17}{c}{\textbf{English}$\rightarrow$\textbf{German}} \\
& \multicolumn{1}{c}{\rotatebox{90}{LIUMCVC\_MNMT\_C}}
& \multicolumn{1}{c}{\rotatebox{90}{UvA-TiCC\_IMAGINATION\_U}}
& \multicolumn{1}{c}{\rotatebox{90}{NICT\_NMTrerank\_C}}
& \multicolumn{1}{c}{\rotatebox{90}{CUNI\_NeuralMonkeyTextualMT\_U}}
& \multicolumn{1}{c}{\rotatebox{90}{DCU-ADAPT\_MultiMT\_C}}
& \multicolumn{1}{c}{\rotatebox{90}{LIUMCVC\_NMT\_C}}
& \multicolumn{1}{c}{\rotatebox{90}{CUNI\_NeuralMonkeyMultimodalMT\_U}}
& \multicolumn{1}{c}{\rotatebox{90}{UvA-TiCC\_IMAGINATION\_C}}
& \multicolumn{1}{c}{\rotatebox{90}{CUNI\_NeuralMonkeyMultimodalMT\_C}}
& \multicolumn{1}{c}{\rotatebox{90}{OREGONSTATE\_2NMT\_C}}
& \multicolumn{1}{c}{\rotatebox{90}{CUNI\_NeuralMonkeyTextualMT\_C}}
& \multicolumn{1}{c}{\rotatebox{90}{OREGONSTATE\_1NMT\_C}}
& \multicolumn{1}{c}{\rotatebox{90}{SHEF\_ShefClassProj\_C}}
& \multicolumn{1}{c}{\rotatebox{90}{SHEF\_ShefClassInitDec\_C}}
& \multicolumn{1}{c}{\rotatebox{90}{BASELINE}}
& \multicolumn{1}{c}{\rotatebox{90}{AFRL-OHIOSTATE\_MULTIMODAL\_C}}\\
LIUMCVC\_MNMT\_C &  & 2.8e-2 & 3.1e-4 & 4.8e-5 & 1.1e-6 & 4.5e-8 & 6.6e-10 & 6.2e-13 & 7.9e-18 & 1.0e-18 & 2.9e-17 & 3.2e-18 & 5.2e-26 & 1.2e-30 & 5.5e-41 & 3.1e-43 \\
\midrule
UvA-TiCC\_IMAGINATION\_U &  & & 4.9e-2 & 1.8e-2 & 1.2e-3 & 6.8e-5 & 3.5e-6 & 2.0e-8 & 3.2e-12 & 2.1e-13 & 4.7e-12 & 5.6e-13 & 3.2e-19 & 9.1e-24 & 3.9e-33 & 1.2e-34 \\
\midrule
NICT\_NMTrerank\_C &  &  &  & - & - & 1.3e-2 & 1.6e-3 & 8.1e-5 & 9.6e-8 & 1.0e-8 & 4.6e-8 & 9.8e-9 & 1.3e-13 & 2.2e-17 & 2.2e-26 & 4.0e-28 \\
CUNI\_NMT\_U &  &  &  &  & - & 4.3e-2 & 1.3e-2 & 6.8e-4 & 2.4e-6 & 4.9e-7 & 1.3e-6 & 3.1e-7 & 1.6e-11 & 6.8e-15 & 3.8e-23 & 4.7e-25 \\
DCU-ADAPT\_MultiMT\_C &  &  &  &  &  & - & 4.0e-2 & 4.9e-3 & 3.0e-5 & 3.3e-6 & 1.3e-5 & 2.2e-6 & 1.4e-10 & 5.3e-14 & 9.8e-23 & 2.4e-24 \\
LIUMCVC\_NMT\_C &  &  &  &  &  &  & - & - & 5.7e-3 & 1.4e-3 & 2.2e-3 & 7.5e-4 & 2.2e-06 & 8.5e-09 & 2.0e-15 & 4.0e-17 \\
CUNI\_MNMT\_U &  &  &  &  &  &  &  & - & 1.7e-2 & 5.7e-3 & 5.6e-3 & 2.8e-3 & 4.5e-06 & 2.7e-08 & 1.2e-15 & 9.7e-18 \\
UvA-TiCC\_IMAGINATION\_C &  &  &  &  &  &  &  &  & - & 2.8e-2 & 3.3e-2 & 1.3e-2 & 6.2e-05 & 3.3e-07 & 3.5e-14 & 5.0e-16 \\
CUNI\_NMMT\_C &  &  &  &  &  &  &  &  &  & - & - & - & 6.5e-03 & 1.5e-04 & 1.5e-10 & 1.7e-12 \\
OREGONSTATE\_2NMT\_C &  &  &  &  &  &  &  &  &  &  & - & - & 1.9e-02 & 1.1e-03 & 2.0e-09 & 3.6e-11 \\
CUNI\_NMT\_C &  &  &  &  &  &  &  &  &  &  &  & - & 4.7e-02 & 3.3e-03 & 8.3e-08 & 2.1e-09 \\
OREGONSTATE\_1NMT\_C &  &  &  &  &  &  &  &  &  &  &  &  & - & 1.2e-02 & 9.1e-07 & 2.6e-08 \\
SHEF\_ShefClassProj\_C &  &  &  &  &  &  &  &  &  &  &  &  &  & - & 4.7e-05 & 1.3e-06 \\
SHEF\_ShefClassInitDec\_C &  &  &  &  &  &  &  &  & & &  &  &  &  & 2.0e-03 & 1.2e-04 \\
\midrule
BASELINE & & & & & & & & & & & & & & &  & - \\
AFRL-OHIOSTATE\_MULTIMODAL\_C & & & & & & & & & & & & & & & & \\
\bottomrule
\end{tabular}
\caption{\label{tab:ende_wilcoxon} English $\rightarrow$ German Wilcoxon signed-rank test at p-level p ≤ 0.05. `-' means that the value is higher than 0.05.}
\end{sidewaystable*}

\begin{sidewaystable*}
\scriptsize
\centering
\begin{tabular}{llllllllllll}
\toprule
\multicolumn{12}{c}{\textbf{English}$\rightarrow$\textbf{French}} \\
& \multicolumn{1}{c}{\rotatebox{90}{NICT\_NMTrerank\_C}}
& \multicolumn{1}{c}{\rotatebox{90}{CUNI\_NeuralMonkeyMultimodalMT\_C}}
& \multicolumn{1}{c}{\rotatebox{90}{DCU-ADAPT\_MultiMT\_C}}
& \multicolumn{1}{c}{\rotatebox{90}{LIUMCVC\_MNMT\_C }}
& \multicolumn{1}{c}{\rotatebox{90}{OREGONSTATE\_2NeuralTranslation\_C}}
& \multicolumn{1}{c}{\rotatebox{90}{CUNI\_NeuralMonkeyTextualMT\_C}}
& \multicolumn{1}{c}{\rotatebox{90}{OREGONSTATE\_1NeuralTranslation\_C}}
& \multicolumn{1}{c}{\rotatebox{90}{LIUMCVC\_NMT\_C}}
& \multicolumn{1}{c}{\rotatebox{90}{SHEF\_ShefClassInitDec\_C}}
& \multicolumn{1}{c}{\rotatebox{90}{SHEF\_ShefClassProj\_C}}
& \multicolumn{1}{c}{\rotatebox{90}{BASELINE}} \\
\midrule
NICT\_NMTrerank\_C &  & - & - & 1.0e-03 & 2.5e-05 & 5.5e-08 & 1.5e-09 & 1.8e-07 & 2.2e-16 & 6.8e-15 & 5.8e-26 \\
CUNI\_NeuralMonkeyMultimodalMT\_C &  & & - & 8.8e-03 & 2.9e-04 & 5.2e-07 & 5.0e-08 & 2.9e-06 & 2.4e-14 & 1.1e-13 & 2.3e-24 \\
DCU-ADAPT\_MultiMT\_C &  &  &  & 2.2e-02 & 1.3e-03 & 6.7e-06 & 7.9e-07 & 2.3e-05 & 2.2e-12 & 1.3e-11 & 3.9e-21 \\
\midrule
LIUMCVC\_MNMT\_C &  &  &  &  & - & 1.7e-02 & 2.7e-03 & 1.2e-02 & 5.3e-07 & 2.0e-06 & 5.9e-14 \\
OREGONSTATE\_2NeuralTranslation\_C &  &  &  &  &  & - & - & - & 1.7e-04 & 1.8e-04 & 7.3e-11 \\
CUNI\_NeuralMonkeyTextualMT\_C &  &  &  &  &  &  & - & - & 7.7e-03 & 7.7e-03 & 5.0e-08 \\
OREGONSTATE\_1NeuralTranslation\_C &  &  &  &  & &  &  & - & 2.4e-02 & 1.8e-02 & 2.3e-07 \\
LIUMCVC\_NMT\_C &  &  &  &  &  &  &  &  & 1.8e-02 & 1.6e-02 & 5.6e-07 \\
\midrule
SHEF\_ShefClassInitDec\_C &  &  &  &  &  &  & &  &  & - & 3.2e-04 \\
SHEF\_ShefClassProj\_C &  &  & &  &  &  &  &  &  &  & 1.8e-03 \\
\midrule
BASELINE & & & & & & & & & & & \\
\bottomrule
\end{tabular}
\captionof{table}{\label{tab:enfr_wilcoxon} English $\rightarrow$ French Wilcoxon signed-rank test at p-level p ≤ 0.05. `-' means that the value is higher than 0.05.}
\end{sidewaystable*}
\newpage
\bibliography{references}
\bibliographystyle{naacl2018}

\end{document}